\documentclass[letterpaper, 10 pt, conference]{ieeeconf}  

\IEEEoverridecommandlockouts                              

\usepackage{amsmath} 

\usepackage{arydshln}
\usepackage{graphics} 
\usepackage{amsmath} 
\usepackage{amssymb}  
\let\labelindent\relax

\usepackage{epsfig}
\usepackage{caption}
\usepackage{lipsum}
\usepackage{multicol}
\usepackage{multirow}
\usepackage{import}
\usepackage{xcolor,colortbl}
\usepackage{enumitem}
\usepackage{cite}
\usepackage{hyperref}
\hypersetup{
    colorlinks=true,
    urlcolor=blue,
}
\usepackage{siunitx}

\usepackage{bbm}

\usepackage[symbol]{footmisc}

\title{\LARGE \bf
Coarse-to-Fine for Sim-to-Real:\\Sub-Millimetre Precision Across Wide Task Spaces
}

\author{Eugene Valassakis$^{1}$, Norman Di Palo$^{1}$, and Edward Johns$^{1}$
\thanks{$^{1}$The Robot Learning Lab at Imperial College London
        {\tt\small eugene.valassakis15@imperial.ac.uk}}%

}
\begin{document}

\maketitle
\thispagestyle{empty}
\pagestyle{empty}

\begin{abstract}
In this paper, we study the problem of zero-shot sim-to-real when the task requires both highly precise control with sub-millimetre error tolerance, and wide task space generalisation. Our framework involves a coarse-to-fine controller, where trajectories begin with classical motion planning using ICP-based pose estimation, and transition to a learned end-to-end controller which maps images to actions and is trained in simulation with domain randomisation. In this way, we achieve precise control whilst also generalising the controller across wide task spaces, and keeping the robustness of vision-based, end-to-end control. Real-world experiments on a range of different tasks show that, by exploiting the best of both worlds, our framework significantly outperforms purely motion planning methods, and purely learning-based methods. Furthermore, we answer a range of questions on best practices for precise sim-to-real transfer, such as how different image sensor modalities and image feature representations perform.
\end{abstract}

\section{INTRODUCTION}

Robot manipulation tasks often require precise control from visual feedback. Traditional methods involve hand-engineered states and controllers, but these require specialist, task-specific design for every new task~\cite{garrett2021integrated, triyonoputro2019quickly}. More recently, end-to-end controllers have addressed this by automatically learning control directly from image observations~\cite{james2017transferring, kalashnikov2018qt}. This involves collecting data with observations and actions, and training a control policy via deep learning, such as through imitation learning or reinforcement learning. However, collecting such data in the real world~\cite{kalashnikov2018qt} can be very time consuming and potentially unsafe.

This can be addressed by using a simulator and zero-shot sim-to-real transfer\cite{james2017transferring,puang2020kovis,valassakis2020crossing}, eliminating the need for any real-world data at all. However, to the best of our knowledge, existing methods in the field either achieve high precision for tasks where the object remains in a very small region of space~\cite{haugaard2020fast}, or they achieve generalisation over wider ranges but do not consider highly precise control~\cite{james2017transferring}. In this paper, as \textbf{our first contribution}, we develop and evaluate a novel \textit{coarse-to-fine} controller for sim-to-real, through which we achieve zero-shot sim-to-real transfer with sub-millimetre (sub-\si{mm}) precision, whilst also generalising across a wide task space, i.e. a wide range of object poses.

In developing this framework, we also found that there are several unanswered questions on what the best practices are for achieving highly precise sim-to-real transfer. For example, existing sim-to-real works consider different input modalities available from cameras, namely depth~\cite{johns2016deep}, stereo IR (often available on depth cameras)~\cite{puang2020kovis}, and RGB~\cite{james2017transferring}, but it is unclear which would work better in an eye-in-hand sim-to-real setting with sub-\si{mm} precision requirements. As another example, image keypoints as a feature representation have been shown to work well~\cite{puang2020kovis}, and conceptually provide a promising, well regularised representation which should facilitate domain-invariance for sim-to-real transfer.
Nonetheless, to the best of our knowledge, this has not yet been verified by comparing keypoints directly to more standard feature representations. Therefore, as \textbf{our second contribution}, we study and evaluate both the sensor modalities and image representations, providing insights on which are most suitable for precise sim-to-real.

\begin{figure}[t!]   
\centering    
\includegraphics[width=0.45\textwidth]{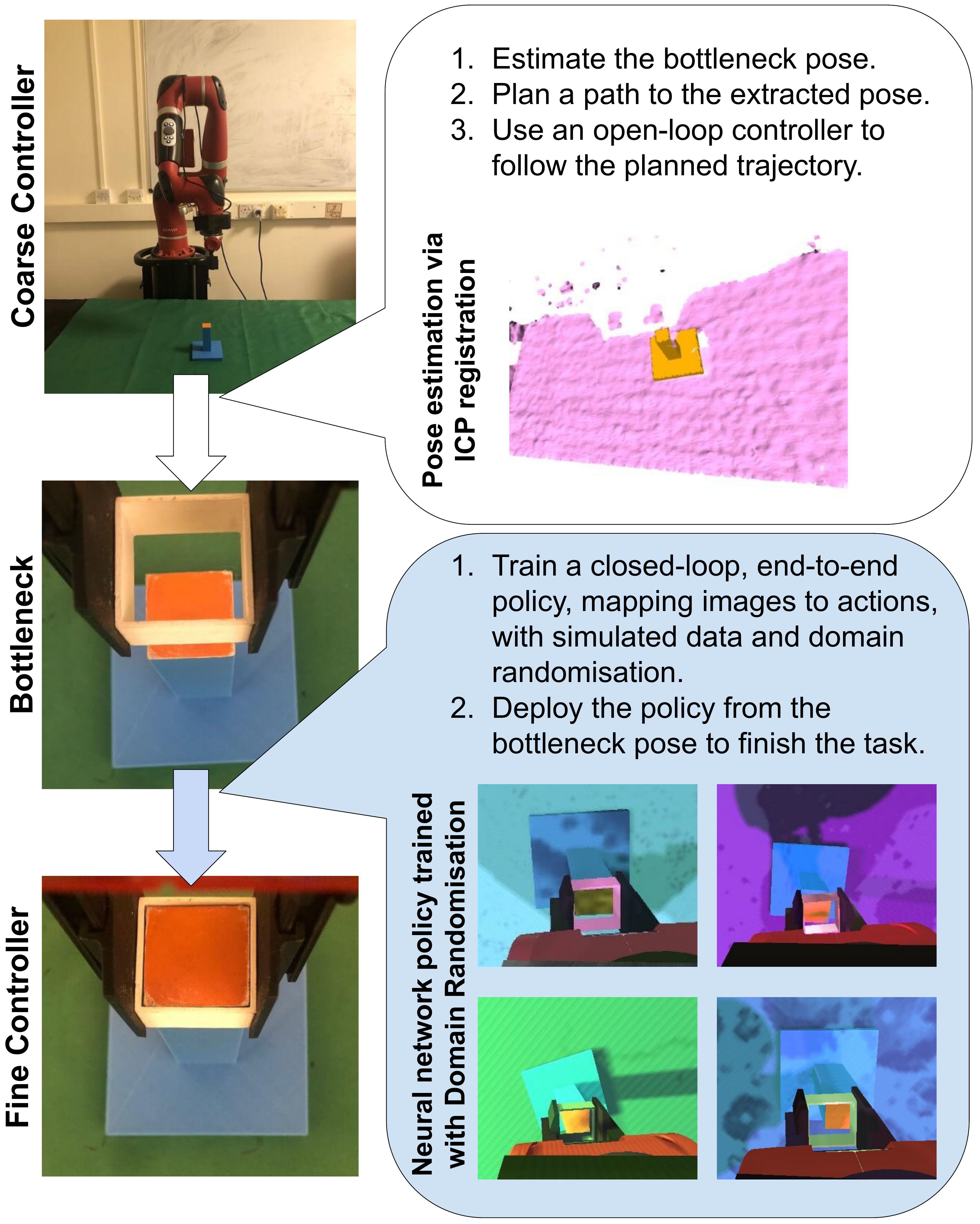}  
\vspace{-0.1cm}
\caption{Overview of our framework.}
\label{fig:framwork}
\vspace{-0.7cm}
\end{figure}

Our \textit{coarse-to-fine for sim-to-real} framework is illustrated in Fig~\ref{fig:framwork}. It consists of a \textit{coarse}, model-based, analytical controller which uses pose estimation to execute broad movements to a \textit{bottleneck} pose, followed by a \textit{fine}, neural network-based, end-to-end controller trained via behavioural cloning in simulation. Suitable for a wide range of applications, we demonstrate its potential by evaluating on three insertion-type tasks with three or four degrees of freedom, satisfying requirements such as wide task space generalisation, precision from $5~\si{mm}$ to sub-\si{mm}, and multi-stage control. We show that our framework is very effective, outperforming both fully model-based and fully learning-based controllers by a large margin. We also found that typical depth sensors today are not likely to yield precise control with an eye-in-hand sim-to-real policy, while RGB inputs alone are sufficient and superior. We also found that, for precise sim-to-real, keypoint-based image features and network architectures seem to yield better results than more standard convolutional architectures. A video and supplementary material for our paper can be found on our website: \url{https://robot-learning.uk/coarse2fine4sim2real}~.

\vspace{-0.1cm}
\section{Related Work}

\subsection{Sim-to-real Transfer for Precise Control}
With the promise of solving the scalability issues associated with using deep learning for control, sim-to-real transfer has been experiencing increasing popularity~\cite{tobin2017domain, james2017transferring, valassakis2020crossing, johns2016deep, andrychowicz2020learning}. Sim-to-real methods typically distinguish between the visual~\cite{alghonaim2021benchmarking} and dynamics~\cite{valassakis2020crossing} aspects of domain transfer, both of which need to be addressed for successfully deploying simulation policies to the real world.

In terms of precise manipulation within the field, Beltran-Hernandez~\cite{beltran2020variable} et al. use sim-to-real training for industrial insertion tasks, tuning the parameters and outputs of an adaptive compliance controller for insertions over a small range. Their work relies on the availability of, possibly noisy, low level state information and force sensing, and is not concerned with processing visual information. Schoettler et al.~\cite{schoettler2020meta} also use low dimensional inputs for industrial insertion tasks, and their work focuses on using few real world trajectories for sim-to-real adaptation via meta-learning.

Using image inputs, Triyonoputro et al.~\cite{triyonoputro2019quickly}, and Haugaard et al.~\cite{haugaard2020fast}, train neural networks on synthetic data and use visual servoing for precise peg-in-hole insertions. Their methods use multiple cameras and predict the pixel position difference / the absolute pixel positions (respectively) of the peg and the hole in image space. They then use that information with hand crafted controllers to move closer and closer to the hole. Both achieve sub-\si{mm} precision, although their controllers are strongly engineered to serve to the last-inch peg-in-hole task without orientation requirements, and do not consider wider or unstructured task spaces.

Closer to our work, Puang et al.~\cite{puang2020kovis}, show that it is possible to be precise with more general, end-to-end sim-to-real policies. Their work however also focuses on last-inch manipulation only, ignoring the issue of how to incorporate such methods to longer horizon tasks. Moreover, the precise M13 bolt and $2~\si{mm}$ error tolerance shaft insertion tasks studied do not seem to require angular alignment to be successfully completed, and use a pre-scripted, open loop controller to finalise the insertion after alignment. As opposed to this, in our work we successfully complete insertions entirely closed-loop, with sub-\si{mm} error tolerance and equally precise angular alignment. We also allow for wide task space generalisation thanks to our coarse-to-fine approach, and provide best practice studies and insights on important hardware and neural network design decisions to accomplish such a task.

\subsection{Coarse-to-fine Controllers for Manipulation}
Robot design encompassing both coarse and fine controllers has been around for many years~\cite{sharon1984enhancement,salcudean1989control}. As such, with the advent of deep learning for control, the idea of combining coarse, model-based controllers with more fine-grained, learning-based methods has naturally emerged as an open research area which has shown some promising results. Johns \cite{johns2021coarse} proposes an imitation learning framework with a coarse controller based on sequential pose estimation, and a fine controller based on behavioural cloning. Lee et al.~\cite{lee2020guided} propose a framework to train a reinforcement learning policy by concentrating training on a region in space where pose estimation uncertainty is too high for a model-based planner. LaGrassa et al.~\cite{lagrassa2020learning} propose a similar idea, but they focus on detecting model failures to execute a model free policy that is learned from demonstrations. Paradis et al.~\cite{paradis2020intermittent} showcase a coarse-to-fine approach for surgical robotics, where they intermittently switch between the coarse and fine controllers for a pick-and-place task. Raj et al.~\cite{raj2020learning} study different schemes for switching neural networks between two modes of operation, namely large and small scale displacements within a given task. In this work we present a coarse-to-fine framework for sim-to-real transfer, and show that we can achieve highly precise manipulation with wide task space generalisation, whilst also only training on simulated data.

\vspace{-0.2cm}
\section{Methods}\label{sec:methods}

\subsection{Coarse-to-fine Framework}\label{sec:framework_method_overview}
The main concept behind our approach is to explicitly distinguish between (1) broad, free space motions that do not require high precision in the control, and (2) highly localised, possibly contact-rich motions that do (see Fig.~\ref{fig:framwork}). This is to disentangle simpler parts of a task which are easily solved with model-based approaches, from harder ones which are better suited to end-to-end learning. In this way, we can ensure that the capacity of the network is reserved for where it is really needed, as opposed to spreading the network's capacity across the entire trajectory.

\subsubsection{\textbf{Coarse Controller}}\label{sec:coarse_traj}Our coarse controller begins at the robot's neutral configuration and ends when the end effector is at a \textit{bottleneck} pose: the point of transition between the coarse and fine controllers. In practice, we define the bottleneck's pose on the object model's frame a few centimetres away from the object, to allow for getting close as possible while still being able to recover from any coarse trajectory positioning errors (see section~\ref{sec:tasks}).  We assume here that the object of interest is within the field of view of a wrist-mounted camera that we use as our vision sensor, and that it is reachable. The former assumption could further be relaxed using a secondary shoulder camera with a broader field of view, but this is left for future work. To complete the coarse trajectory, we leverage the power of model-based motion planners and classical controllers in order to generalise across the task space. As inputs to the motion planner, we require models for the robot, the object and the outside environment:

We first obtain the pose of the object of interest in the robot's frame. We do so using a point cloud from a depth sensor and Iterative Closest Point (ICP)~\cite{paul1992besl}. ICP requires an initialisation matrix, which we obtain using a neural network. This network uses an RGB image to predict (1) the $u,v$ pixel coordinates of a pre-defined, fixed point on the object, (2) the depth value of that pixel, and (3) the rotation angle of the object frame around the vertical direction. Using these and the intrinsic and extrinsic camera matrices we obtain an initial estimate the object's pose in the robot frame. This prediction does not need to be very accurate, simply enough to be an effective initialisation for ICP. To remain in the zero-shot sim-to-real regime we train this network entirely with simulated data, as described in  section~\ref{sec:data_gen}. Once the object's pose is obtained, we use the pose of the bottleneck in the object model's frame to obtain a final target pose for the end effector. We can then find and execute a trajectory to the bottleneck. In our experiments we craft a proportional controller in end effector space and Inverse Kinematics (IK) to follow a linear path to the bottleneck pose, although any motion planner could be used here instead.

\begin{figure}[t]   
\centering    
\includegraphics[width=0.45\textwidth]{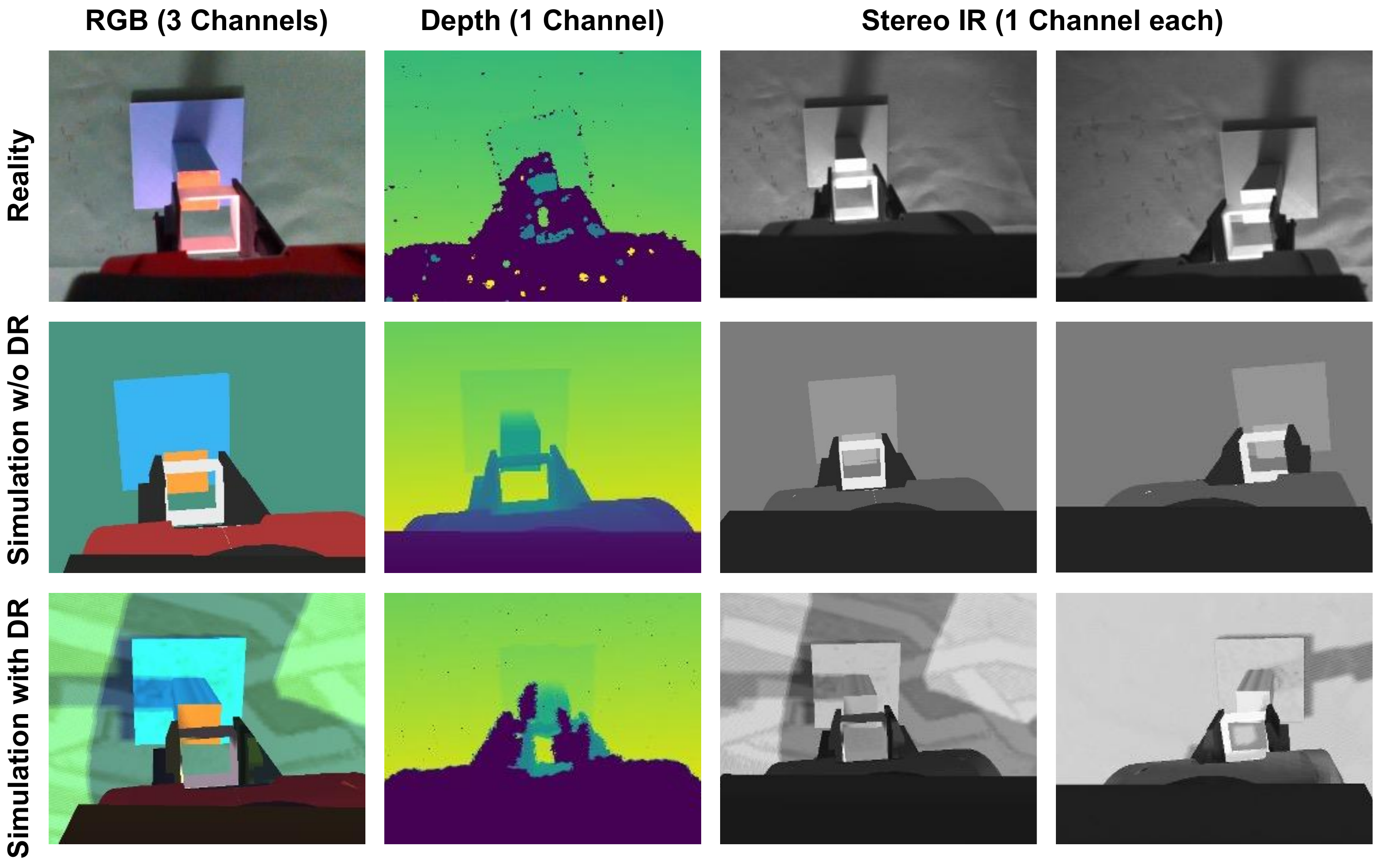}  
\vspace{-0.2cm}
\caption{Examples of real-world, and randomised/non-randomised simulation images from different sensors.}
\label{fig:randomisaiton_examples}
\vspace{-0.7cm}
\end{figure}

\subsubsection{\textbf{Fine Controller}}\label{sec:fine_overview}If we were able to obtain perfect pose estimations from the coarse controller, then most tasks could be solved with simple analytical controllers and motion planners. However, inaccuracies in the camera calibration, available models, depth sensors, and ICP optimisation can accumulate, causing the coarse controller to fail where precise control is necessary. To solve this while maintaining generality, we deploy a fine-grained, end-to-end neural network controller. This controller begins when the coarse one has positioned the robot at the bottleneck pose, and runs closed-loop all the way until the task is successfully completed (see Fig~\ref{fig:framwork}). The network receives image inputs from the visual sensors and outputs velocity commands in the end effector frame, which are then converted to joint velocity targets through IK. 

Since the camera is now localised close to the target object and the commands are in the end effector frame, the network has no need to be globally aware of the task space. As such, we gain the ability to crop the image tightly around the end effector, maintaining the object of interest in view. This allows us to keep higher resolution inputs for fine-grained control while keeping the network sizes reasonable, as well as naturally filtering out any background distractors. Overall, these benefits allow the networks to focus their capacity on achieving zero-shot sim-to-real transfer for this small region of space, making it possible to solve very precise manipulation tasks, despite the reality gap.

\subsection{Networks and Training}\label{sec:networks_and_training}
\vspace{-0.1cm}
A major benefit of using simulations for training is that they give us access to perfect knowledge of the state of the world, allowing us to craft expert policies with relative ease. We exploit this privilege and use behavioural cloning to train our sim-to-real policies. As such, our networks map high-dimensional image inputs  $\mathbf{o}$ (see Fig.~\ref{fig:randomisaiton_examples}),  to low dimensional actions $\mathbf{y}$ for control. In our experiments, these consist of the Cartesian velocities $v_x,v_y,v_z$ and the angular velocity $\alpha$ around the $z$ axis, all with respect to the end effector frame. 

As our main model we use a spatial encoder-decoder keypoint architecture \cite{finn2016deep}. Specifically, we encode the input images into $K$ $2D$ keypoints, $\{ (u_i,v_i) \}_{i=1}^K$, before decoding them to (1) the actions, and (2) a depth reconstruction auxiliary output. Each keypoint represents a pair of coordinates in image space, and can autonomously learn to track a particular point in the scene~\cite{finn2016deep}.  The depth reconstruction output is added to help the network understand the geometric properties of the scene via an auxiliary loss during training, as well as offering a level of interpretability to the network's behaviour (see Fig~\ref{fig:qualitative}). We note that this step is only possible thanks to the privileged information accessible from simulators, namely perfect depth maps from each sensor.

Formally, we first use an encoder network parameterised by $\theta$ to map the raw observations into keypoints,  $e_{\theta}:\mathbf{o} \in \mathbb{R}^{M\times N\times C}\rightarrow \mathbf{k}\in \mathbb{R}^{K \times 2}$, where  $K$ is the number of keypoints,  and $M \times N \times C$ the input image  dimension. Specifically, we pass the input into a succession of convolutional layers with BatchNorm~\cite{ioffe2015batch} and ReLU~\cite{nair2010rectified} activations, which results in a lower dimensional feature map  $\mathbf{h} \in \mathbb{R}^{I\times J\times K}$. We then transform this feature map into the set of keypoints by doing a spatial soft-argmax operation \cite{finn2016deep}: First, a 2D channel-wise softmax operation is applied:
\begin{equation}
    o_{i,j,k} = \frac{e^{h_{i,j,k}}}{\sum_{i,j}e^{h_{i,j,k}}},
\end{equation}
where $0 \leq i \leq I,~0 \leq j \leq J$ are the spatial indices of the $k$-th channel of the feature maps. Then, a soft-argmax is applied to both the $u$ and $v$ directions to obtain the keypoint $(u_k,v_k)$ by a weighted sum:
\begin{equation}
    u_{k} = \frac{1}{I}\sum_i^I i \sum_j^J o_{i,j,k} \text{,   } v_{k} = \frac{1}{J}\sum_j^J j \sum_i^I o_{i,j,k}.
\end{equation}

These keypoints are then passed through our policy network parameterised by $\phi$, $f_{\phi}:\mathbf{k} \in \mathbb{R}^{K\times2}\rightarrow \mathbf{y}\in \mathbb{R}^{n}$, with $n=4$ in our experiments. The network is a multi-layer perceptron (MLP) with BatchNorm, Dropout \cite{srivastava2014dropout} and ReLU activations on each hidden layer. The final layer is a simple linear map, allowing us to regress the velocity command $\textbf{y}$.

We also pass the keypoints through an auxiliary branch, which reconstructs the depth values at each pixel position of the original image,  from the view of the input camera sensor, $d_\zeta:\mathbf{k}\in \mathbb{R}^{K\times2}\rightarrow \mathbf{d}\in\mathbb{R}^{M\times N\times1}$: We first transform each keypoint $(u_i, v_i)$ into a $2D$ Heatmap~\cite{jakab2018unsupervised} $\mathbb{\mathbf{M}}^i$,

\begin{equation}
  [\mathbb{\mathbf{M}}^i]_{m,n} \propto \mathcal{L}\left(m | u_i, \mathbf{\sigma}\right) \times\mathcal{L}\left(n | v_i, \mathbf{\sigma}\right),
\end{equation}

with $\mathcal{L}$ the Laplace distribution centred at the keypoint and with scale  $\mathbf{\sigma}$ fixed for all maps. We then concatenate the $K$ maps channel-wise, and pass them through a series of upscaling convolutions with BatchNorm and ReLU activations, to obtain the depth reconstruction $\mathbf{d}$.

During training, we obtain the perfect depth map $\mathbf{d}^*$ and expert actions $\mathbf{y}^*$ (see section~\ref{sec:tasks}), and use an L2 loss to minimize the reconstruction error from the depth maps $l_{\text{rec}}=||\mathbf{d}-\mathbf{d}^*||_2$, and the sum of an L2 and L1 losses to optimise the actions $l_{\text{actions}} =||\mathbf{y}-\mathbf{y}^*||_2+||\mathbf{y}-\mathbf{y}^*||_1 $, with our total loss being a combination of the two : $L=l_{\text{rec}}+l_{\text{actions}}$. Finally, we note that the architecture and training methods are the same in the case of the ICP initialisation network, with the only difference being that the final outputs are the predictions described in section~\ref{sec:coarse_traj}. Detailed hyperparameter values and network diagrams can be found in our supplementary material.


\begin{table}[t]
\caption{Summary of the randomised simulation aspects for dataset generation.\label{table:randparams}}
\begin{tabular}{l|l}
                        & \textbf{Randomised Simulation Aspects}                                                               \\ \hline
\textbf{Visual Sensing} & \begin{tabular}[c]{@{}l@{}}Ambient light, main light source, secondary  light  \\ source, colours, textures, depth noise, depth  masks \end{tabular} \\ \hline
\textbf{Geometry}       & \begin{tabular}[c]{@{}l@{}}Vision sensor pose, initial scene configuration, \\ gripper component poses, image crop centres \end{tabular}                                       \\ \hline
\textbf{Dynamics}       & \begin{tabular}[c]{@{}l@{}}Random action noise and expert policy diversity \end{tabular}     
\end{tabular}
\vspace{-0.5cm}
\end{table}

\subsection{Data Generation}\label{sec:data_gen}

\subsubsection{\textbf{Overview}} We generate all our data in a simulator, which we set up as described in section~\ref{sec:experimental_setup}.  For each task, we craft appropriate expert policies, described in~\ref{sec:tasks}, and use them to collect demonstrations in simulation.  For the ICP initialisation data, we sample different initial positions of the object of interest, and record the corresponding images and ground truths for the predictions described in~\ref{sec:coarse_traj}.

\subsubsection{\textbf{Overcoming the Reality Gap}} Operating in the zero-shot setting  creates the problem of the ``reality gap'', which we overcome by using domain randomisation~\cite{tobin2017domain, valassakis2020crossing}. Table~\ref{table:randparams} summarises which aspects of the simulation we randomise to account for visual sensing, dynamics, and geometry, and a full breakdown with exact parameters and  value ranges can be found in our supplementary material. Visual randomisation tackles discrepancies in colours, textures, and lighting. In order to account for these, we (1) randomise the colour of each relevant simulation component around a mean extracted from several real images (for the object/table), or obtained from coloured mesh files (for the robot), (2) randomise the light source, light colours, and ambient light properties, and (3) add random grayscale textures to each relevant component of the simulation.  Dynamics randomisation accounts for hard-to-model physical processes, and discrepancies in physical parameters. As shown in~\cite{valassakis2020crossing}, an effective method for accounting for this is to simply inject noise into the simulation state though random forces applied to the relevant components. In our case we follow a similar principle and create diversity in the visited simulation states by adding random noise at each commanded velocity~\cite{andrychowicz2020learning}, as well as adding diversity in the paths followed by the expert policies that are collecting demonstrations (see section~\ref{sec:tasks}). Finally, we account for geometry misalignments between the simulation and the real world by randomising the pose of the camera, gripper components and grasped objects at each timestep, as well as the cropping positions for each image.

\subsubsection{\textbf{Simulating Depth Images}}For experiments where assisted stereo depth images are used as inputs to the networks, we have also accounted for the reality gap by (1) zeroing out unrealistic random artefacts appearing in real depth images, and (2) providing the appropriate randomisations (see Fig.~\ref{fig:randomisaiton_examples}). For the latter, we build upon the procedure described in~\cite{thalhammer2019sydd}. One of the major issues with assisted stereo which does not occur in simulations, is missing depth values due to unobservable IR data in the scene caused by a baseline between the emitter and sensor(s). In order to render depth images that reflect this, we first simulate the pattern emitter of the depth camera with a light source. We then capture this light on the two sensors representing the stereo cameras, which results into two binary masks indicating the pixels where emitter light is visible. Inverting these two masks, projecting them onto the depth image and keeping the union leaves us with a binary mask of occlusions where the depth image should have 0 value. In order to increase its realism and add randomisation, this occlusion mask is further processed with morphological opening, followed by morphological dilation and median filtering, all with randomised kernel sizes~\cite{thalhammer2019sydd}.

In order to get the final depth images, we combine the perfect depth image with this occlusion mask and further apply the following steps: (1) We augment the  depth values with proportional noise according to the real camera's depth profile~\cite{ahn2019analysis}, (2) We set a cutoff minimum depth, (3) We warp the depth image with perlin noise, and apply Gaussian filtering~\cite{zakharov2018keep,thalhammer2019sydd}, and (4) We randomly set $0.1\%$ of depth pixels to 0. Several of these operations rely on parameters which we randomise, and a detailed list can be found in our supplementary material.

\section{Experiments}\label{sec:experiments}
Our experiments are designed to show the effectiveness of our proposed framework, as well as as study the effect of different design choices when setting up a sim-to-real training pipeline. In this section, we start by giving an overview of our experimental setup in~\ref{sec:experimental_setup}, followed by a description of our tasks in~\ref{sec:tasks}. We then evaluate our overall framework under standard conditions in~\ref{sec:exp_overall}, and test it under challenging ones in~\ref{sec:exp_robustness}. Finally, in~\ref{sec:exp_input_mod} and~\ref{sec:exp_network_arch} we study the effect of the input sensor modality and image representation on achieving higly precise sim-to-real transfer.

\subsection{Experimental Setup}\label{sec:experimental_setup}
For all our experiments we used the Sawyer robot, which we fitted with a Realsense D435 depth camera using a $3D$-printed wrist mount. Our experimental setup can be seen in Fig.~\ref{fig:framwork}. The Sawyer's neutral position places the end effector at about $28~\si{cm}$ above the table, and our task space is roughly $30\times 35~\si{cm}$, but with an irregular shape since we are ensuring the peg to be visible from the camera. For our simulator, we use CoppeliaSim~\cite{coppeliaSim} with PyRep~\cite{james2019pyrep}. To set up our simulation, we use the Sawyer URDF and CAD models, the D435 specifications, CAD models of our test objects and camera mount, an estimate of the camera extrinsics obtained from calibration, and real-world measurements of the table. We set our camera resolution to $848\times480$, and generate our policy inputs by cropping each image to a $256\times256$ window centered around the end effector, before resizing it to $64\times64$. For the  ICP initialisation networks the images are not cropped, simply resized to $128\times128$ with padding where necessary to avoid distortions.

\begin{figure}[t!]   
\centering    
\includegraphics[width=0.45\textwidth]{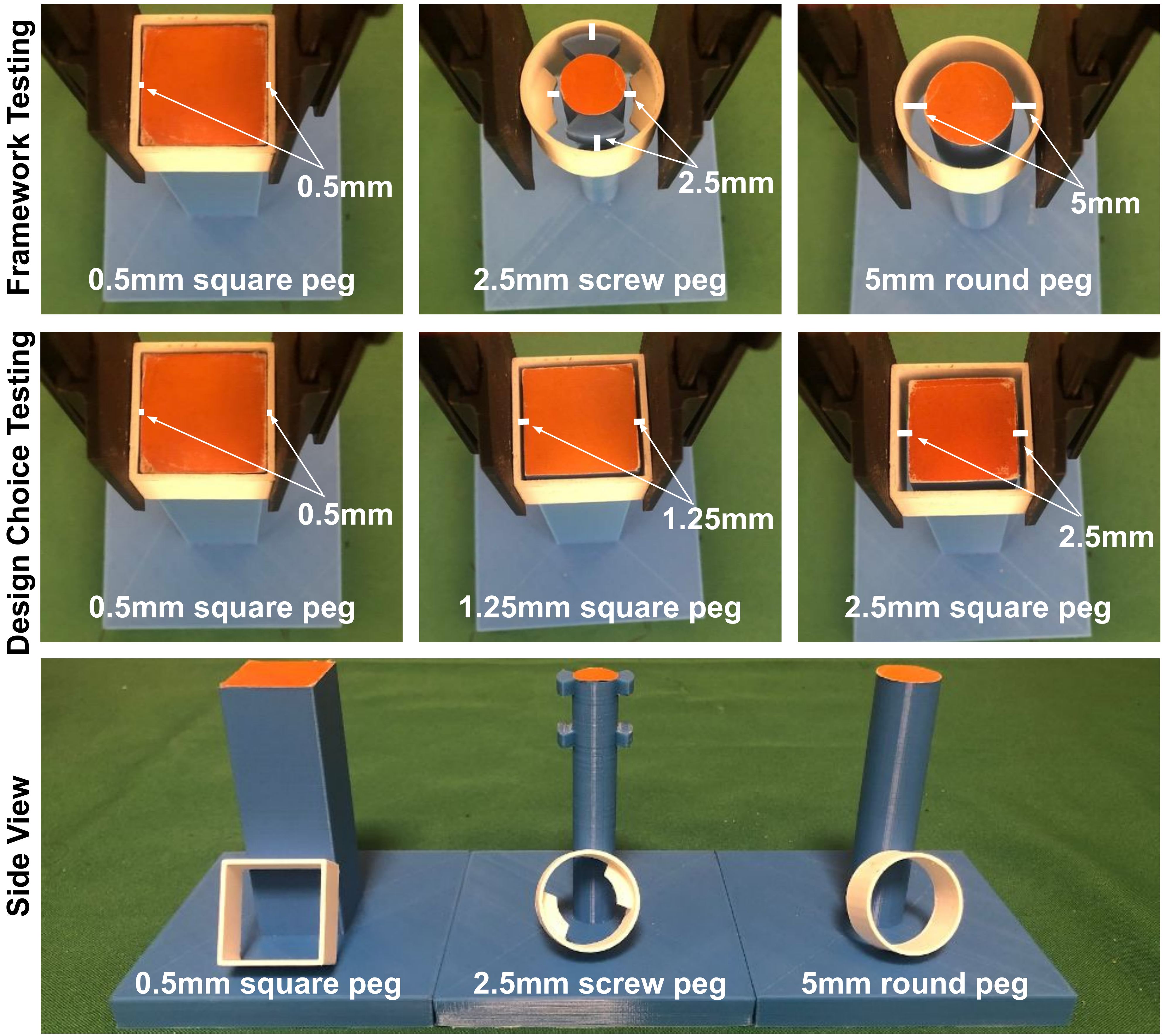}  
\vspace{-0.1cm}
\caption{Illustration of the pegs used for our tasks.}
\label{fig:pegs}
\vspace{-0.7cm}
\end{figure}

\vspace{-0.2cm}
\subsection{Tasks}\label{sec:tasks}
In this section, we detail the three different tasks we use for our various experiments, which are illustrated in Fig.~\ref{fig:pegs}. They consist of three different types of insertion that span various difficulties, and allow us to explore very fine-grained manipulation, wide task space generalisation, and multi-stage policies. For each task/difficulty level, we gather data in simulation and train policies separately. To gather the data, we start by sampling an initial pose centred at the  \textit{bottleneck}. The bottleneck poses are defined $3~\si{cm}$ above each of our pegs, such that a straight downward motion would result in insertion. Both the initial position and orientation are sampled uniformly, within a $2.5~\si{cm}\times2.5~\si{cm}\times1.5~\si{cm}$ rectangular volume and  $[-0.45,0.45]~rad$ range, respectively. 

We then roll out hand crafted expert policies that use privileged simulation information in order to collect demonstrations. At each simulation timestep $t$, we record the visual observation $\mathbf{o_{t,i}}$ from each type of vision sensor $i$, the optimal action $\mathbf{y^*_t}$, and a perfect depth image from each sensor, $\mathbf{d^*_{t,i}}$. Detailed implementation descriptions of our expert policies can be found in our supplementary material.

\subsubsection{\textbf{Square Peg Insertion}}
The goal is to insert a square ring over a square peg. To complete the task, our controller needs to align the ring in position and orientation and then complete the insertion by lowering it onto the peg.

\subsubsection{\textbf{Insertion with Screw Motion}} The goal is to insert an irregular shaped ring onto an irregular shaped peg (see Fig.~\ref{fig:pegs}), and then twist it  in order to ``secure'' the insertion. The task is successful if a vertical upwards motion would not remove the ring from the peg, and a vertical downward motion would also get blocked. It tests the robustness of our framework's close-loop, end-to-end fine controller to more complex, multi-stage control requirements. 

\begin{figure*}[thb!]   
\centering    
\includegraphics[width=\linewidth]{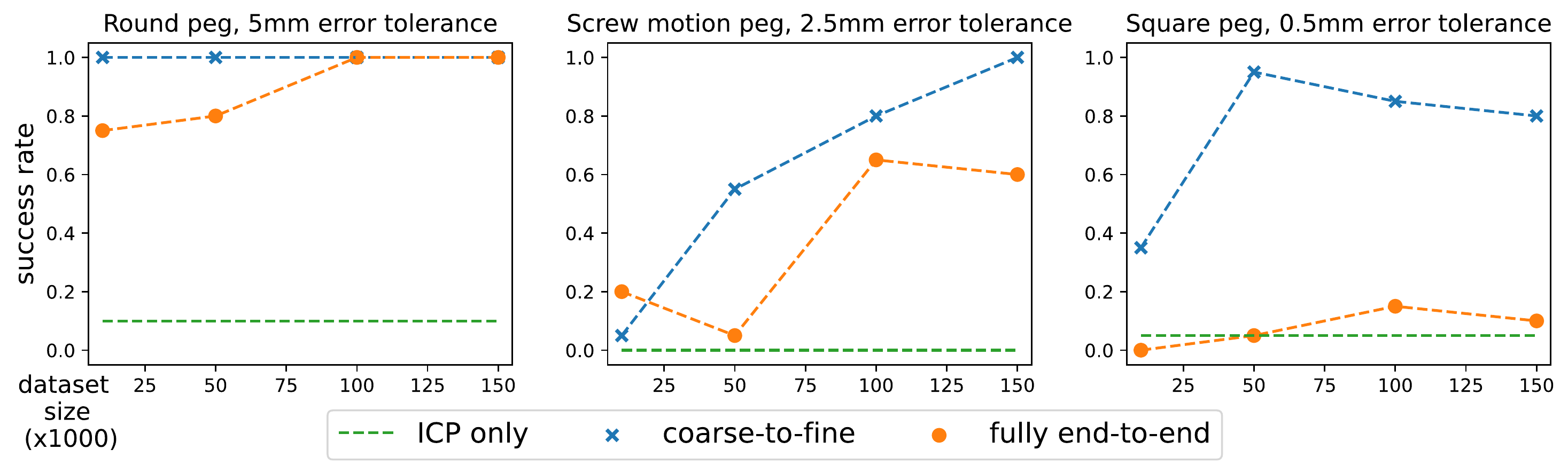}  
\vspace{-0.6cm}
\caption{Success rate vs dataset size for our three different tasks.}
\label{fig:main_results}
\vspace{-0.5cm}
\end{figure*}

\subsubsection{\textbf{Round Peg Insertion}}
The simplest of our tasks, which is successful upon inserting a cylindrical ring over a cylindrical peg.  As such, the only relevant control dimensions are the $x,y,z$ linear velocities, and the task only requires positional alignment to be completed.

\subsection{Overall Framework Evaluation}\label{sec:exp_overall}
\subsubsection{\textbf{Experimental Procedure}} In this experiment we aimed to test our overall coarse-to-fine framework. We used RGB inputs, which we found work best (see section~\ref{sec:exp_input_mod}). For testing, we used a $5~\si{mm}$ error tolerance round peg insertion, a $2.5~\si{mm}$ tolerance screw motion peg insertion, and a $0.5~\si{mm}$ tolerance  square peg insertion. These tolerances correspond to the maximum deviation from the centre of the peg that would still result in a successful insertion, and are illustrated in the first row of Fig.~\ref{fig:pegs}. We further designated $20$ initial peg poses on the table, where the peg is visible by the camera from the robot's neural position, and is easily reachable given the robot kinematics. For each of those we recorded whether the corresponding trajectory succeeded on the task, from which we calculated a success rate.

We compared our method with two baselines. First, we used an \textit{ICP-only} baseline where the coarse controller remains the same, but at the bottleneck we repeated the pose estimation using a second depth image and the ICP algorithm. Using this, we created a controller to complete the task analytically, without the use of any neural network, and followed it in an open-loop fashion. Second, we used a \textit{fully end-to-end} policy which uses a neural network in order to perform the entire trajectory all the way from the neutral position of the robot, without any model-based components in the controller. In training this policy we used the same network architecture and the data collection procedure with domain randomisation described in~\ref{sec:methods}, with the exceptions that (1) the expert policies were adapted to accommodate large displacements in a reasonable amount of time, and (2) the entire $848\times480$ images, padded and resized to $128\times128$, were used as inputs to keep the object always in view.

\begin{figure}[t!]   
\centering    
\includegraphics[width=0.45\textwidth]{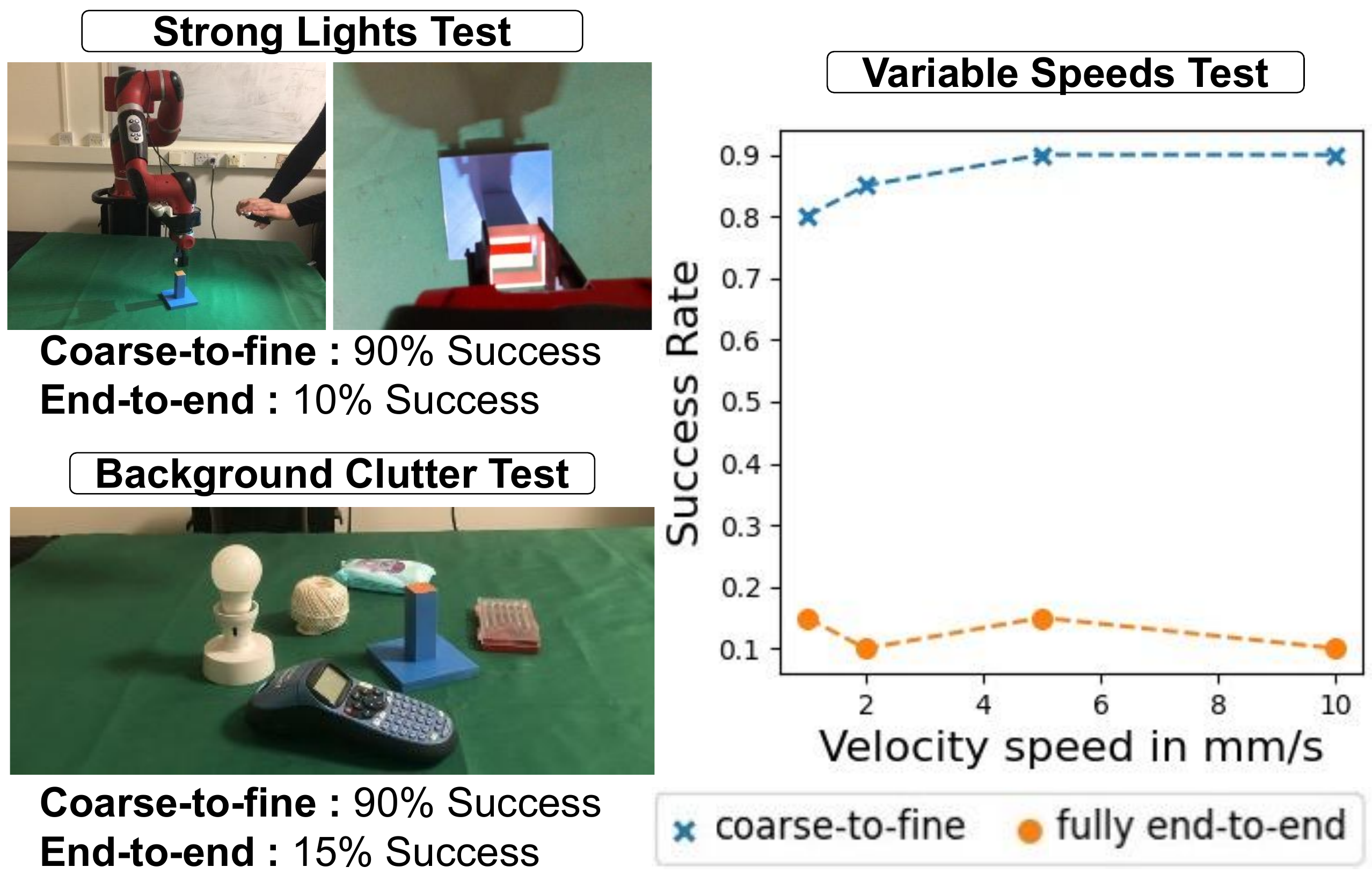}  
\vspace{-0.1cm}
\caption{Robustness and Stress Test Results.}
\label{fig:robustness_test}
\vspace{-0.7cm}
\end{figure}

\subsubsection{\textbf{Results}}
We repeated this experiment for the coarse-to-fine and end-to-end policies trained with $\{10,50,100,150\}\times1000$ datapoints, and show the results in Fig.~\ref{fig:main_results}. We note that since the ICP-only method does not contain any trained policy, it only produces one set of results. Across all tasks, we see that using our coarse-to-fine approach largely outperforms the baselines, with a performance that reaches close to or at 100\%. We also note that our policies quickly scale with the dataset size, then seemingly showing a slight decrease in performance after a certain point, most notably in the $0.5~\si{mm}$ square insertion task. We believe this is because after a certain size the dataset is sufficient to accommodate the task, at which point noisy sources of variation in the network training and setup start having noticeable effects.

\subsection{Robustness and Stress Testing}\label{sec:exp_robustness}
\subsubsection{\textbf{Experimental Procedure}}
In this experiment we aim to stress test our method under challenging conditions. Specifically, we answer the following questions: (1) Does our method generalise well to hard light conditions and background distractors, and (2) is it applicable with variable speeds to allow for customisable, both fast and accurate manipulation? In order to answer these, we used our best performing policy with  RGB inputs on the $0.5~\si{mm}$ error tolerance square insertion task, and for completeness we also evaluated the performance of its fully end-to-end counterpart under the same conditions. As before, we measured the success rate over the same $20$ initial peg poses.  In order to evaluate (1), we first used a torchlight that we waved around the setup during insertion and second added random objects around the table. For (2), we simply scaled the velocity of the policies to different fixed speeds during deployment, when the manipulator reached near the bottleneck, namely for the fine part of the trajectory. 

\subsubsection{\textbf{Results}}
 Our results for both these experiments can be seen in Fig.~\ref{fig:robustness_test}. We can see that even under challenging conditions such as strong lighting or background clutter, our networks generalise well, which we attribute to the effectiveness of our domain randomisation procedure. We additionally see that executing either our method or the fully end-to-end controller at different speeds does not seem to significantly affect the success rate, even when using operating speeds well outside the training range. This is useful as it shows our method can also accommodate variable task execution time requirements.

\begin{figure*}[t!]   
\centering    
\includegraphics[width=\linewidth]{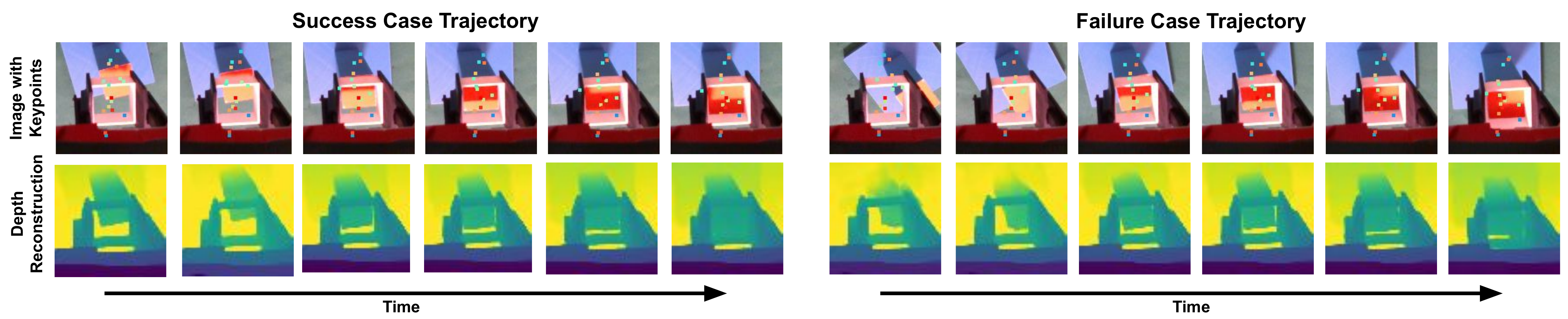}  
\vspace{-0.6cm}
\caption{Illustration of a success and failure case trajectories on the $0.5~\si{mm}$ square peg task.   }\label{fig:qualitative}
\vspace{-0.5cm}
\end{figure*}


\subsection{Deciding on an Input Modality}\label{sec:exp_input_mod}
\subsubsection{\textbf{Experimental Procedure}}
When training a policy for sim-to-real manipulation with high-dimensional visual sensor inputs, one has the choice between several sensing modalities to use. In a typical assisted stereo depth camera, such as the D435 we used in our experiments, we can for example readily query RGB, depth and IR images (see Fig.~\ref{fig:randomisaiton_examples}). Past sim-to-real works have chosen between those in order to construct their methods~\cite{puang2020kovis,james2017transferring,johns2016deep}, but it is unclear to the best of our knowledge if there is any merit in choosing any one of those modalities over the other. In this experiment, we answer this question by benchmarking the following input options: (1) RGB, (2) stereo IR, (3) depth, and (4) grayscale. Examples of those can be seen in Fig~\ref{fig:randomisaiton_examples}, except for grayscale which corresponds to the RGB images simply converted to a 1-channel grayscale image. Since we are interested here in how the input modalities affect the policy controller specifically, we focused this experiment on the fine control part of the pipeline. In order to ensure a fair comparison, we kept the same network architecture for each experiment, with the main difference being the number of channels on the network inputs. For stereo IR, images were concatenated channel-wise. We note that in early experiments, we also considered propagating each IR image through independent networks, but we did not notice significant difference in performance, hence we opted to keep the same architecture. 

For a granular comparison, we used for this experiment the square peg task only, but considered $0.5~\si{mm}, 1.25~\si{mm}$ and $2.5~\si{mm}$ error tolerances (see second row in Fig.~\ref{fig:pegs}). We proceeded by placing the end effector at the bottleneck position (see Fig.~\ref{fig:framwork}), followed by sampling $20$ initial poses around it. Each is obtained by sampling uniformly a $\pm1.5~\si{cm}$ positional error in each direction and a $\pm15^\circ$ error in the angle around the vertical axis. Once sampled, these were recorded and remained fixed across all the policies tested. For each initial position, we deployed the policy that is being tested, and recorded whether the trajectory resulted in a successful insertion.

\begin{table}[t]
\small
\setlength{\tabcolsep}{1.5pt}
\centering
\caption{Comparison of the success rate achieved on the square peg task from the different input sensor modalities. \label{table:input_mod}}
\addtolength{\leftskip} {-0.1cm} 
\begin{tabular}{c|c|c|c}
                       & \textbf{2.5~mm tolerance} & \textbf{1.25~mm tolerance} & \textbf{0.5~mm tolerance} \\ \hline
\textbf{Depth}         & 0.8                    & 0.3                      & 0                      \\         
\textbf{IR}            & 1                      & 1                        & 0.05                   \\
\textbf{Grayscale} & 1                      & 1                        & 0.55    \\
\textbf{RGB}           & 1                      & 1                        & 1                  

\end{tabular}
\vspace{-0.6cm}
\end{table}

\vspace{-0.1cm}

\subsubsection{\textbf{Results}}
Table~\ref{table:input_mod}, shows the performance achieved by the different available input modalities. We see that the RGB, stereo IR and grayscale networks perform equally well for error tolerances as small as $1.25~\si{mm}$, while the depth-based network trails behind. This is not surprising looking at Fig~\ref{fig:randomisaiton_examples}: the real depth images were not sufficiently accurate at those ranges of operation, and it seems that even our domain randomisation procedure was not enough to create highly performing policies on those inputs. Looking at the $0.5~\si{mm}$ error tolerance results, we see that the RGB-based network maintains its performance, while there is a clear decrease for its stereo IR and grayscale counterparts. This is somewhat surprising since conceptually grayscale images should contain sufficient information for completing the insertions, while stereo IR images should be even more informative for inferring depth positioning. We conjecture that possible explanations are (1) it is easier to saturate IR pixel values, leading to loss of information, (2) in grayscale the boundaries between the peg and the ring may be harder to detect, and (3) the information redundancy on the RGB channels may help. Nonetheless, we note that it is also possible that hard to control factors, such as the particular sensor positions, may be playing a role, or that further engineering of our domain randomisation and network parameters may have compensated for the differences. At the very least, we can draw the interesting  conclusion that using a single RGB sensor is sufficient for achieving sim-to-real transfer for manipulation with sub-\si{mm} precision requirements.

\begin{table}[t]
\small
\setlength{\tabcolsep}{1.5pt}
\centering
\caption{Comparison of the success rate achieved on the square peg task from the different architectures.\label{table:archit}}
\addtolength{\leftskip} {-0.1cm} 
\begin{tabular}{c|c|c|c}
                                                                           & \textbf{\begin{tabular}[c]{@{}c@{}}2.5~mm \\ tolerance \end{tabular}} & \textbf{\begin{tabular}[c]{@{}c@{}}1.25~mm \\ tolerance \end{tabular}} & \textbf{\begin{tabular}[c]{@{}c@{}}0.5~mm \\ tolerance \end{tabular}} \\ \hline
\textbf{\begin{tabular}[c]{@{}c@{}} Convolutions\end{tabular}}                                                               &          1           &       0.95            &0.55  \\

\textbf{\begin{tabular}[c]{@{}c@{}} Convolutions \& Depth Recon.\end{tabular}} & 1                      & 1                        & 0.6                      \\
\textbf{\begin{tabular}[c]{@{}c@{}}Keypoints \end{tabular}} & 1                      & 1                        & 1                      \\ 
\textbf{\begin{tabular}[c]{@{}c@{}}Keypoints \&  Depth Recon. \end{tabular}} & 1                      & 1                        & 1                      
\end{tabular}
\vspace{-0.5cm}
\end{table}

\subsection{Deciding on a Feature Representation}\label{sec:exp_network_arch}
\subsubsection{\textbf{Experimental Procedure}}
In order for an image-based neural network to transfer well between a simulator and the real world, it is important that its learned features are invariant to the visual discrepancies between the two domains. Traditionally, domain randomisation with simple convolutional features have been used~\cite{james2017transferring}, and recently encoder-decoder structures with keypoint representations have also shown good results~\cite{puang2020kovis}. Conceptually, the latter should provide strong regularisation, helping the network focus on the geometric aspects of the image, especially when coupled to a depth reconstruction auxiliary loss. This in turn should help to better filter out the irrelevant sources of variation in domain randomised images, ultimately resulting in better sim-to-real transfer. This experiment is aimed to verifying this intuition experimentally under a controlled environment.

In order to do so, we again used the square peg task with $\{2.5~\si{mm},1.25~\si{mm},0.5~\si{mm}\}$ error tolerances, and the same initial poses as the input modality experiment. We then deployed four different architectures, and recorded the resulting success rates: (1) our keypoint-based encoder-decoder architecture using the auxiliairy depth reconstruction loss, (2) our keypoint-based architecture without using the depth reconstruction loss, (3) an encoder-decoder network with regular convolutional features and an auxiliairy depth reconstruction loss, and (4) a convolutional features encoder without any depth reconstruction. All architectures used RGB image inputs, and we also kept their number of parameters as close as possible (within roughly $5\%$ variation).

\subsubsection{\textbf{Results}}
We summarise the results of this experiment in Table~\ref{table:archit}. Looking at the table, it becomes apparent that that the addition of the keypoints as a feature representation results in a significant jump in performance. This seems consistent with the intuition that keypoints offer a better regularisation for the feature space, helping the policies develop robustness to the reality gap. On the other hand, adding an encoder-decoder structure with a depth reconstruction auxiliary loss did not seem to have any significant impact on the performance, regardless on whether keypoints were used. Nonetheless, we argue that it is still recommended to use a depth reconstruction branch. The reason becomes apparent when looking at Fig.~\ref{fig:qualitative}, which shows several images across a successful trajectory and a failure case. In the top row we show the keypoints that are predicted for each image, and in the bottom row the depth reconstructions that result. As we can see, having access to such information adds a layer of intepretability to the network's behaviour, which not only greatly facilitates the development process but could also be used to assess the network's confidence during deployment, and potentially catch hazardous states. For instance, in Fig.~\ref{fig:qualitative} we can clearly see that in the failure case the depth reconstructions are much worse than in the success case, and in fact on the last image we can see that the network believes there is still a gap between the ring and the peg, while in fact there is none.

\vspace{-0.1cm}
\section{Conclusions}
In this work, we have presented our framework for zero-shot sim-to-real transfer of control policies with sub-\si{mm} precision and wide task space generalisation capabilities. It combines a classical, model-based controller using pose estimation to execute a coarse trajectory to a bottleneck pose, and a learning-based, end-to-end policy which starts at the bottleneck and completes the task in a closed-loop manner. In real-world experiments, we found that our framework achieves sub-\si{mm} precision across a wide task space, whilst maintaining performance under challenging conditions. We also found that for precise sim-to-real, RGB network inputs perform better than both depth and stereo IR, and that network architectures with keypoint-based image representations result in better policies than more standard convolutional architectures.
\vspace{-0.2cm}

\addtolength{\textheight}{-12cm}   






\bibliography{references}
\bibliographystyle{abbrv}

\end{document}